\documentclass[conference]{IEEEtran}
\IEEEoverridecommandlockouts
\usepackage{cite}
\usepackage{amsmath,amssymb,amsfonts}
\usepackage{algorithmic}
\usepackage{graphicx}
\usepackage{textcomp}
\usepackage{xcolor}
\usepackage{booktabs}
\usepackage{multicol}
\usepackage{multirow}

\def\BibTeX{{\rm B\kern-.05em{\sc i\kern-.025em b}\kern-.08em
    T\kern-.1667em\lower.7ex\hbox{E}\kern-.125emX}}
\begin{document}

\title{\fontsize{22}{28}\selectfont Exploring Vision-Language Models for Online Signature Verification: A Zero-Shot Capability Study}

\author{
Marta Robledo-Moreno, Ruben Vera-Rodriguez, Ruben Tolosana, Javier Ortega-Garcia\\
\normalsize \textit{BiometricsAI, School of Engineering, Universidad Autonoma de Madrid, Spain}\\
{\tt\small \{marta.robledo, ruben.vera, ruben.tolosana, javier.ortega\}@uam.es} \\
}


\maketitle

\begin{abstract}
Recent advancements in Vision-Language Models (VLMs) have demonstrated strong capabilities in general visual reasoning, yet their applicability to rigorous biometric tasks remains unexplored. This work presents an exploratory study evaluating the zero-shot performance of state-of-the-art VLMs (GPT-5.2 and Gemini 2.5 Pro) on the Signature Verification Challenge (SVC) benchmark. To enable visual processing, raw kinematic time-series are converted into static images, encoding pressure information into stroke opacity whenever available in the source data. Furthermore, we introduce a scoring protocol that extracts latent token probabilities to compute robust biometric scores. Experimental results reveal a significant performance dichotomy dependent on signal quality and forgery type. In random forgery scenarios, the zero-shot VLM achieves exceptional discrimination, with GPT-5.2 reaching an Equal Error Rate of 0.32\% in mobile tasks, outperforming supervised state-of-the-art systems. Conversely, in skilled forgery scenarios, where the task is more challenging because both signatures are almost identical, the results are significantly worse, and a critical ``Rationalization Trap" emerges: chain-of-thought (CoT) reasoning degrades performance as the model produces kinematic hallucinations to justify forgery artifacts as natural variability.
\end{abstract}


\section{Introduction}
\label{sec:intro}
Behavioral biometrics refer to a form of authentication that involves the analysis of distinctive patterns exhibited by a user's activity to verify their identity \cite{stylios_behavioral}. This includes traits such as signature verification~\cite{tolo_deepsign}, in-air signature verification~\cite{airsign}, keystroke dynamics~\cite{nahuel_type}, or gait recognition~\cite{paula_gait}.  In particular, online signature verification (confirming the authenticity and integrity of a signature) is widely deployed in banking, legal, and security sectors due to its capacity to capture dynamic time-series data (including coordinates or pressure on the screen) rather than mere static shapes. The state of the art (SOTA) in this domain is dominated by highly specialized narrow AI architectures, such as Time-Aligned Recurrent Neural Networks (TA-RNNs) \cite{tolo_deepsign} or Deep Soft-DTW \cite{jiang_dsdtw}, introduced by Tolosana \textit{et al.} and Jiang \textit{et al.}, respectively. While these systems achieve remarkable performance, they suffer from the inherent limitations of narrow AI: they require extensive training on massive domain-specific datasets and lack the flexibility to adapt to new acquisition scenarios (e.g., switching from stylus to finger inputs) without architectural reconfiguration.

\begin{figure}[t]
    \centering
    \includegraphics[width=\columnwidth]{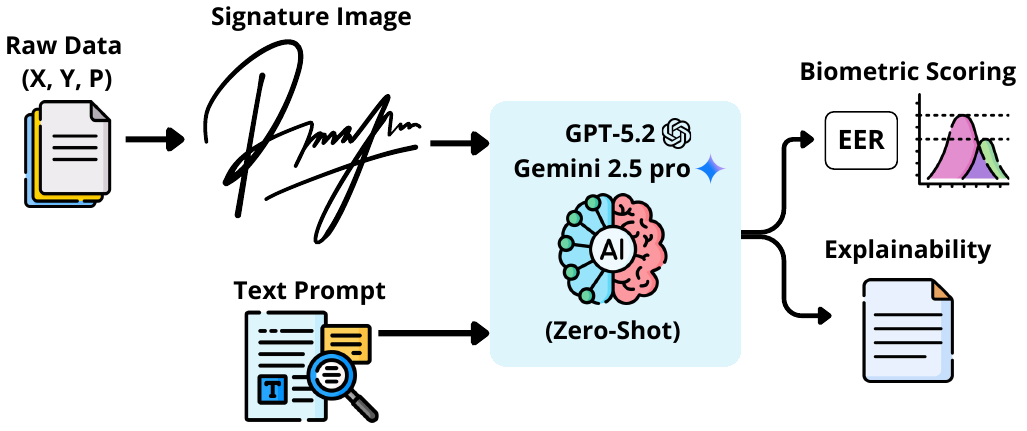}
    \caption{\textbf{Overview of the Proposed Zero-Shot Forensic Framework.} Raw kinematic sensor data (blind to VLMs) is transformed into static representation. A General-Purpose VLM, prompted as a Forensic Document Examiner, analyzes the image. We extract both Log-Likelihood Scores for robust biometric evaluation and Textual Rationales for post-hoc explainability, auditing the model's reliability boundaries.}
    \label{fig:graphical_abstract}
\end{figure}

\vspace{3mm}
Recently, the field of Artificial Intelligence has experienced a significant transformation due to the emergence of General-Purpose AI Systems (GPAIS) \cite{triguero_general}. The development of Large Language Models (LLMs) has evolved into interfaces capable of performing in a wide range of tasks that extend beyond their initial training objectives.

The integration of visual encoders with pre-trained LLMs has established a new paradigm in deep learning: Vision-Language Models (VLMs) \cite{bordes_vlms}. By bridging the gap between high-dimensional visual data and discrete linguistic concepts, architectures ranging from contrastive frameworks like CLIP~\cite{radford_clip} to generative models~\cite{gen} have demonstrated remarkable zero-shot competence across diverse domains \cite{thirunavukarasu_large}~\cite{adish_visual}~\cite{Enkelejda_chatedu}. These systems, often trained on massive web-scale datasets, have evolved beyond simple classification to perform complex reasoning tasks. However, applying these general-purpose models to forensic biometrics introduces severe reliability challenges that remain unaddressed. Research indicates that VLMs frequently suffer from hallucinations \cite{bai_hallucination}, producing content that is neither required nor relevant to the input. Furthermore, while standard VLM benchmarks focus on semantic understanding or spatial reasoning, signature verification demands the discrimination of fine-grained kinematic artifacts to which static visual encoders may be blind. This raises fundamental questions for the community: \textit{Can VLMs, largely trained on natural images, be trusted to verify whether two signatures belong to the same person without specific fine-tuning?} And critically, \textit{can their discrete text-generation capabilities be adapted to yield the rigorous continuous probabilities required for standard biometric evaluation?}

\vspace{9mm}

In this work, we propose a novel zero-shot framework to audit the readiness of VLMs for high-precision forensic biometrics. We hypothesize that by employing intelligent preprocessing to adapt dynamic kinematic signals into pressure-encoded visual representations, and by using the Log-Likelihood scoring mechanism inherent to Large Language Models, forensic verification can be performed competently. This approach allows us to rigorously test the application boundaries of foundation models without the need for task-specific fine-tuning.  All resources related to this study are available for the research community\footnote{https://github.com/BiDAlab/VLMs-SigVer}.

\noindent Our main contributions are:
\begin{itemize}
    \item[-] We provide the first analysis of two of the main commercial state-of-the-art VLMs for online signature verification, assessing whether a general-purpose system can compete with specialized narrow AI baselines without any training in the Signature Verification Challenge \cite{tolo_svc}.
    \item[-] We explore the extraction of token \textit{log-probabilities} to obtain superior biometric performance compared to textual confidence scores, providing a robust mechanism to evaluate generalist models with high-precision metrics.
    \item[-] We explore the VLM's Chain-of-Thought capabilities to provide interpretable forensic reports, analyzing the coherence between the model's reasoning and its final verification verdict.
\end{itemize}

The remainder of the paper is organized as follows: Section \ref{sec:related} reviews related work on signature verification and VLMs. Section \ref{sec:methodology} details the proposed zero-shot framework. Section \ref{sec:results} presents the experimental benchmarking on the Signature Verification Challenge (SVC) dataset, followed by an explainability exploration in Section \ref{sec:explain}. Section \ref{sec:conclusion} draws conclusions and future research directions.
\section{Related Work}
\label{sec:related}
This section contextualizes our work within the evolution of signature verification, contrasting traditional deep learning approaches with the emerging capabilities of VLMs in handwriting understanding and reasoning.

\vspace{-1.5mm}
\subsection{Online Signature Verification}
For the past decade, online signature verification has been the domain of highly specialized narrow AI. State-of-the-art systems rely on architectures explicitly designed to model time-series dynamics, such as Time-Aligned Recurrent Neural Networks \cite{tolo_deepsign}, which are based on a first stage where different time functions are extracted from the original signals and a second one where they are aligned and processed through a Recurrent Neural Network. Another approach is Deep Soft-DTW \cite{jiang_dsdtw}, where developers proposed a method in which they use neural networks to learn deep time functions as inputs for Dynamic Time Warping (DTW). This enables end-to-end learning of optimal alignment paths, allowing the model to automatically compensate for non-linear temporal distortions, such as fluctuations in signing speed, that differ between genuine samples and forgeries.

However, a critical limitation of these SOTA architectures is their lack of transparency. As defined in the taxonomy of Explainable AI (XAI) by Arrieta \textit{et al.} \cite{arrieta_xai}, deep learning models typically operate as ``black-box" systems, offering high performance but obscuring the reasoning behind their decisions. This lack of interpretability is a major barrier for Responsible AI in high-stakes forensic scenarios, where a verdict must be accompanied by a justifiable rationale. In this context, VLMs offer an alternative path: specific forms of post-hoc explainability. VLMs can generate natural language explanations, bridging the cognitive gap between the AI's decision and the human expert. In this work, we examine not only the biometric accuracy of these models but also the coherence of their textual reasoning, assessing whether they can serve as trustworthy agents within the XAI framework.

\vspace{-1.2mm}
\subsection{VLMs in Handwriting}
The emergence of VLMs has introduced zero-shot competence in domains previously requiring supervised training. In the context of handwritten documents, recent benchmarks have demonstrated that VLMs are evolving from passive observers to active reasoners.
Díez \textit{et al.} \cite{diez_evaluating} evaluated VLMs for handwritten text recognition, showing that general-purpose models can transcribe historical and contemporary manuscripts with competitive accuracy, effectively understanding the content of the stroke. 
Beyond transcription, Nath \textit{et al.} \cite{nath_can} introduced the FERMAT benchmark to assess whether these models can grade handwritten mathematics. Their findings suggest that VLMs possess the capacity to detect errors and reason over spatial perturbations in handwritten digits.

However, current VLMs have not been rigorously tested in the forensic task of determining \textit{``who"} wrote it. This task requires analyzing subtle kinematic dynamics and topological features, a capability that extends beyond simple pattern recognition.

\subsection{Exploring Zero-Shot Biometric Capabilities}
Applying VLMs to biometric verification presents unique challenges, primarily the gap between high-level semantic understanding and low-level metric precision. DeAndres-Tame \textit{et al.} \cite{deandres_chatgpt} pioneered this exploration in the domain of face biometrics, benchmarking ChatGPT-4 against specialized feature extractors like ArcFace. Their study revealed that while VLMs demonstrate competence in soft biometrics (distinguishing subjects based on semantic attributes such as gender, ethnicity, or accessories), they struggle significantly with identity verification. Their work introduced the methodology of extracting self-reported certainty scores from the model's text output to construct Detection Error Tradeoff (DET) curves, a protocol we adapt in this study. However, they notably cautioned that the model's explainable textual rationales often suffer from calibration errors, where the VLM expresses high confidence in incorrect verifications. This precedent highlights the necessity of moving beyond purely textual metrics to audit the underlying probabilistic reliability of general-purpose agents in forensic tasks.

\vspace{-1.5mm}

\section{Methodology}
\label{sec:methodology}

To audit the readiness of VLMs for forensic biometrics, we designed a zero-shot experimental framework that adapts raw sensor data into a visual modality and extracts granular biometric scores from the model's latent space. This section details the dataset protocol, the preprocessing pipeline, and the multi-stage scoring mechanisms employed.

\subsection{Database and Protocol}
We utilized the evaluation dataset of the Signature Verification Challenge \cite{tolo_svc}, a complex dataset containing dynamic signatures acquired under varying conditions. The evaluation is structured into three specific tasks to test the model's adaptability boundaries:
\begin{itemize}
    \item[-] Task 1 (Office - Stylus): Analysis of signatures captured with a digital pen on a specialized device, containing pressure information (6,000 comparisons).
    \item[-] Task 2 (Mobile - Finger): Analysis of mobile signatures performed with a finger (9,520 comparisons).
    \item[-] Task 3 (Combined): A mixed scenario evaluating cross-device robustness (12,000 comparisons).
\end{itemize}

\subsection{Signature Preprocessing}
Since standard VLMs are blind to raw time-series text files, we first transform raw data to static images. For Task 1, we propose a pressure-encoded representation. We map the normalized pressure signal $P(t)$ to a gray scale, where high-pressure segments are rendered as darker strokes. This encoding allows the VLM to visually infer different profiles. For Task 2, where pressure is undefined, the algorithm adapts by rendering uniform binary strokes, testing the model's ability to perform pure geometrical verification.

\subsection{Model Specifications and Configuration}
In this work, we tested two state-of-the-art multimodal foundation models, configuring them with strict parameters to favor determinism. For the primary agent, we selected GPT-5.2, OpenAI's latest model\footnote{https://openai.com/index/introducing-gpt-5-2/}. The API interaction was configured with \texttt{temperature=0.0} and a fixed seed (\texttt{seed=42}) to minimize stochasticity, and crucially, we enabled \texttt{logprobs=True} to access the raw logits of the verification tokens, allowing us to compute more accurate biometric scores required for the Equal Error Rate (EER) analysis. As a secondary agent, we employed Gemini-2.5-Pro, Google DeepMind's advanced reasoning model\footnote{https://blog.google/innovation-and-ai/models-and-research/google-deepmind/gemini-model-thinking-updates-march-2025/}, known for its thinking process updates. Due to the strict safety policies of commercial VLM providers regarding the processing of human biometric data, raw signature samples often trigger automated privacy filters that prevent inference. To ensure a reproducible evaluation pipeline, and inspired by DeAndres-Tame \textit{et al.} \cite{deandres_chatgpt}, a task-neutral clarification was included in the prompt labeling the images as ``AI-generated". It is important to emphasize that this labeling is a technical requirement for API compatibility and does not provide any semantic or class-specific information regarding the authenticity of the signature, thus maintaining the integrity of the zero-shot experimental protocol.

\subsection{Two-Stage Reasoning}

\begin{figure}[t]
    \centering
    \includegraphics[width=\columnwidth]{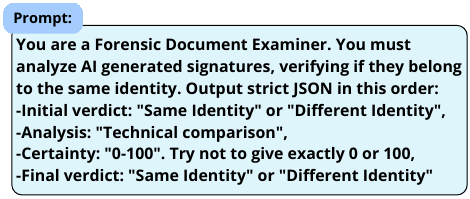} 
    
    \caption{\textbf{System prompt} utilized to instantiate the VLM as a Forensic Document Examiner. The structure is proposed to capture the transition from a reflexive Initial Verdict to a Final Verdict via a CoT phase.}
    \label{fig:prompt}
    \vspace{-3mm}
\end{figure}

We designed a specific prompt strategy to capture the evolution of the model's decision-making process (see Figure~\ref{fig:prompt}). The model, acting as a ``Forensic Document Examiner", is instructed to output a strict JSON containing two distinct verdicts: a first decision based on the immediate visual impression of the signatures and a reflective decision reached after the explicit reasoning step. We also ask the model for the certainty of the final verdict. This structure serves a twofold purpose. First, it allows us to measure whether the reasoning process reinforces correct decisions or induces hallucinations in the final verdict. Second, it enables a direct comparison between the model's self-reported textual certainty and the raw log-probabilities. By restricting the output to ``Same Identity" or ``Different Identity", we were able to extract the specific probabilities of these tokens (in GPT-5.2) and convert them into a continuous biometric score.

\subsection{Biometric Scoring}
A critical challenge in using generative models for biometrics is converting qualitative text into quantitative scores for DET curve calculation. In this work, we compute three distinct scores to evaluate the model's performance:

\subsubsection{Log-Likelihood Extraction ($S_{v1}$ and $S_{v2}$)}
To bridge the gap between discrete text generation and continuous biometric scoring, we use the \texttt{logprobs} parameter exposed by the VLM inference API. Given that our system prompt enforces a strict structural constraint (outputting only ``Same Identity" or ``Different Identity"), we adopt an assumption presuming that the probability mass is almost exclusively concentrated between these two token classes.

Instead of performing a full vocabulary softmax, we implement a direct extraction strategy based on the top-1 generated token $t$. The biometric score $S_{vx}$ (representing the probability of a genuine match) is derived as follows:

\begin{equation}
    S_{vx} = 
    \begin{cases} 
    \exp(\mathcal{L}_t) & \text{if } t \in T_{same} \\
    1 - \exp(\mathcal{L}_t) & \text{if } t \in T_{diff}
    \end{cases}
\end{equation}

\noindent where $\mathcal{L}_t$ is the log-probability of the generated token, and $T_{same}$ / $T_{diff}$ represent the sets of tokens associated with positive (e.g., ``Same") and negative (e.g., ``Different") verdicts, respectively. This formulation efficiently maps the VLM's confidence into a normalized $[0,1]$ similarity space suitable for EER calculation. Specifically, $S_{v1}$ (Initial Impression) is derived from the \texttt{initial\_verdict} token, capturing the model's immediate visual intuition prior to explicit reasoning, while $S_{v2}$ (Reflective Decision) is derived from the \texttt{final\_verdict} token, quantifying the confidence level established after the Chain of Thought (CoT) analysis.

\subsubsection{Self-Reported Certainty Score ($S_{text}$)}
To audit the semantic calibration, we extract the explicit certainty value (0-100) requested in the system prompt. This raw integer is first normalized to a unit scale $[0,1]$. However, unlike the probabilistic score which inherently represents the likelihood of a ``Match", the textual confidence indicates certainty in the specific chosen verdict. To ensure metric consistency, we apply a directional alignment: if the model predicts ``Different Identity", the normalized confidence is inverted ($S_{text} = 1 - \text{confidence}$). This transformation ensures that $S_{text}$ operates in the same similarity space as $S_{vx}$ (where $1.0$ implies a perfect genuine match).
\vspace{2mm}

\section{Experimental Results}
\label{sec:results}

We evaluated the zero-shot verification performance of the proposed framework across the three SVC-onGoing \cite{tolo_svc} tasks. Results are reported in terms of EER. To contextualize the performance gap between generalist agents and specialized systems, we compare our approach against the participants of the challenge, whose systems are described in the original work by Tolosana \textit{et al.} \cite{tolo_svc}.

Table \ref{tab:results_comparison} presents the consolidated metrics to ensure a fair comparison with the published benchmark. To understand the reliability boundaries of the model, in Table \ref{tab:detailed_results} we break down performance into two distinct biometric scenarios:
\begin{itemize}
    \item[-] Random forgeries: Comparisons involving two signatures from two different identities (different name/morphology).
    \item[-] Skilled forgeries: Comparisons where an impostor explicitly attempts to imitate the genuine user's signature (similar shape, different dynamics).
\end{itemize}

\subsection{General-Purpose vs. Specialized Systems}

Table \ref{tab:results_comparison} benchmarks the proposed framework against the full SVC leaderboard. As expected, the top-tier supervised systems (e.g., DLVC-Lab), trained specifically on dynamic signals, achieve the lowest error rates. However, a striking finding emerges in Task 2 (Mobile Finger Scenario): our zero-shot GPT approach ($9.49\%$) not only surpasses the standard SVC Baseline (DTW, $14.92\%$) but also outperforms several specialized supervised submissions. We hypothesize that this occurs because finger-based signatures are kinematically noisy, which can confound rigid mathematical distance metrics and even specialized feature extractors. In contrast, the VLM, operating in the visual domain, captures the global shape topology, effectively ignoring the high-frequency signal noise that degrades traditional algorithms. 
It is noteworthy that the explicit reasoning phase significantly enhances GPT-5.2's performance in Tasks 2 and 3.  In the context of the mobile scenario, the EER improves from $18.79\%$ (initial impression, $S_{v1}$) to $9.49\%$ (reflective CoT, $S_{v2}$). A similar trend is observed in the combined scenario, where the error rate decreases from $23.17\%$ to $16.95\%$.

A closer inspection of the two VLM architectures reveals a distinct performance dependent on signal quality. In the high-fidelity Task 1, both models perform comparably, with Gemini 2.5 Pro achieving a slight edge ($22.90\%$ vs. $24.62\%$). However, this advantage disappears in the noisy Task 2. Here, GPT-5.2 demonstrates significantly superior performance, achieving an EER of $9.49\%$ compared to Gemini's $16.94\%$. In Task 3, the same behavior is reported (GPT achieves a 16.95\% compared to Gemini's $24.67\%$.

\begin{table*}[t]
    \centering
    \caption{\textbf{Benchmarking Results (EER \% $\downarrow$) on the SVC evaluation dataset \cite{tolo_svc}.} Comprehensive comparison of the proposed Zero-Shot VLMs framework against the full leaderboard of the SVC Competition. The upper section lists state-of-the-art supervised systems, while the lower section presents our zero-shot approach.}
    \label{tab:results_comparison}
    \setlength{\tabcolsep}{10pt} 
    \begin{tabular}{l l c c c}
        \toprule
        \textbf{System Type} & \textbf{Method / Team} & \textbf{Task 1} & \textbf{Task 2} & \textbf{Task 3} \\
         & & \textit{(Stylus - Office)} & \textit{(Finger - Mobile)} & \textit{(Combined)} \\
        \midrule
        \multirow{8}{*}{\textit{Supervised (SVC-onGoing)}} 
         & DLVC-Lab (Winner) & 3.33 & 7.41 & 6.04 \\
         & BiDA-Lab & 4.08 & 8.67 & 7.63 \\
         & TUSUR KIBEVS & 6.44 & 13.39 & 11.42 \\
         & SIG & 7.50 & 10.14 & 9.96 \\
         & MaD & 9.83 & 17.23 & 14.21 \\
         & SigStat & 11.75 & 13.29 & 14.48 \\
         & SVC-Baseline (DTW) & 13.08 & 14.92 & 14.67 \\
         & JAIRG & -- & 18.43 & -- \\
        \midrule
        \midrule
        \multirow{4}{*}{\textit{Zero-Shot (Ours)}} 
         & GPT-5.2 ($S_{v1}$) & 24.62 & 18.79 & 23.17 \\
         & GPT-5.2 ($S_{v2}$) & 24.91 & \textbf{9.49} & \textbf{16.95} \\
         & GPT-5.2 ($S_{text}$) & 24.62 & 19.26 & 19.26 \\
         \cmidrule{2-5}
         & Gemini 2.5 Pro ($S_{text}$) & \textbf{22.90} & 16.94 & 24.67 \\
        \bottomrule
    \end{tabular}
    \vspace{2mm}
    \\
    \footnotesize{\textit{Note: $S_{v1}$ denotes Initial Impression Score; $S_{v2}$ denotes Reflective CoT Score; $S_{text}$ denotes Self-Reported Certainty. Supervised results from \cite{tolo_svc}.}}
\end{table*}

\subsection{Skilled vs. Random Forgeries}
In random scenarios, both VLMs show exceptional discrimination. Notably, GPT ($S_{v2}$) achieves an EER of 0.32\% in Task 2, surpassing not only the standard DTW baseline (3.30\%) but also the supervised SOTA (SIG: 3.62\%). This indicates that for morphological verification, zero-shot visual reasoning is already superior to specialized time-series algorithms trained on the specific dataset. The Chain-of-Thought successfully filters these easy negatives, driving the error rate near zero.

In skilled scenarios, results bifurcate based on signal quality. In the high-fidelity stylus task, the VLM fails to match the precision of specialized deep learning systems (DLVC-Lab: 4.66\%), achieving $\sim 35\%$ of EER. Here, the Rationalization Trap \cite{sevastjanova_rationalizationtrap} is most evident: explicit reasoning ($S_{v2}$) degrades performance from $35.16\%$ to $48.08\%$ (near random guessing). Qualitative analysis suggests that when faced with high-quality forgeries, the model hallucinates justifications to rationalize the visual similarity, becoming confident in a wrong decision. However, a critical inflection point appears in Task 2. Here, GPT ($S_{v2}$) achieves $18.64\%$, overtaking the SVC-Baseline (DTW: $20.58\%$).

\begin{table}[t!]
    \centering
    \caption{\textbf{Zero-Shot vs. Reference (EER \% $\downarrow$).} Detailed performance breakdown across Random and Skilled scenarios, comparing the proposed VLM framework against SVC baselines \cite{tolo_svc}.}
    \label{tab:detailed_results}
    
    \resizebox{\columnwidth}{!}{%
    \begin{tabular}{lllccc}
        \toprule
        \multirow{2}{*}{\textbf{Scenario}} & \multirow{2}{*}{\textbf{System}} & \multirow{2}{*}{\textbf{Metric}} & \textbf{Task 1} & \textbf{Task 2} & \textbf{Task 3} \\
         & & & \textit{(Stylus)} & \textit{(Finger)} & \textit{(Combined)} \\
        \midrule
        
        \multirow{6}{*}{\textbf{Random}} 
          & \multirow{3}{*}{GPT (Ours)} 
          & $S_{v1}$ (Initial) & 4.57 & 6.86 & 5.92 \\
          & & $S_{v2}$ (Reasoning) & \textbf{0.49} & \textbf{0.32} & \textbf{0.60} \\
          & & $S_{text}$ (Certainty) & 4.26 & 2.86 & 3.96 \\
          \cmidrule{2-6} 
          & Gemini (Ours) & $S_{text}$ (Certainty) & 5.58 & 3.41 & 4.62 \\
          \cmidrule{2-6}
          & \textit{SIG} & \textit{Supervised (SOTA)} & 1.08 & 3.62 & 2.50 \\
          & \textit{SVC-Baseline} & \textit{DTW} & 1.00 & 3.30 & 2.58 \\
        
        \midrule
        
        \multirow{6}{*}{\textbf{Skilled}} 
          & \multirow{3}{*}{GPT (Ours)} 
          & $S_{v1}$ (Initial) & 35.16 & 27.69 & 33.14 \\
          & & $S_{v2}$ (Reasoning) & 48.08 & 18.64 & 32.20 \\
          & & $S_{text}$ (Certainty) & 35.82 & 37.68 & 32.74 \\
          \cmidrule{2-6}
          & Gemini (Ours) & $S_{text}$ (Certainty) & 35.35 & 28.34 & 33.69 \\
          \cmidrule{2-6}
          & \textit{DLVC-Lab} & \textit{Supervised (SOTA)} & \textbf{4.66} & \textbf{8.77} & \textbf{6.79} \\
          & \textit{SVC-Baseline} & \textit{DTW} & 16.83 & 20.58 & 19.45 \\
        
        \bottomrule
    \end{tabular}%
    }
\end{table}
\section{Explainability and Reliability}
\label{sec:explain}
Beyond quantitative accuracy, the adoption of AI in security depends on trust: as stated in \cite{mancera_5pba}, the integration of LLMs in high-stakes applications raises critical transparency, privacy, and ethical demands. Following the XAI taxonomy \cite{arrieta_xai}, our VLM-based system falls under the category of Post hoc textual explainability, where the model generates a rationale to justify its prediction. In this section, we show a few examples of generated CoT to explore this capability (see Figure \ref{fig:qualitative_analysis}).

\subsection{Random Forgeries (Topological Reasoning)}
\label{subsec:random_analysis}

In random forgery scenarios (bottom row of Figure \ref{fig:qualitative_analysis}), the forensic task relies primarily on morphological discrimination. Our exploration reveals that both GPT-5.2 and Gemini 2.5 Pro excel in this regime, effectively functioning as geometrical reasoners. Visually, the signatures present clear structural disparities: the genuine sample is segmented, while the impostor sample is continuous. The generated CoT reflects a high degree of fidelity to these visual ground truths. GPT's analysis accurately characterizes the genuine signature as being composed of ``localized tall verticals", contrasting it with the impostor's ``strong horizontal extension". Similarly, Gemini provides a sophisticated geometric abstraction, defining the conflict as a ``bipartite signature" versus a ``single, integrated unit" with ``elliptical flourishes". In this scenario, the textual explainability is highly reliable. The models identify fundamental differences that objectively exist in the image. This semantic accuracy correlates perfectly with the probabilistic metrics (Certainty $ \geq 86\%$), providing forensic examiners with valid, verifiable rationales.

\subsection{Skilled Forgeries and the ``Rationalization Trap"}
\label{subsec:skilled_analysis}

\begin{figure*}[h!]
    \centering
    \includegraphics[width=0.99\textwidth]{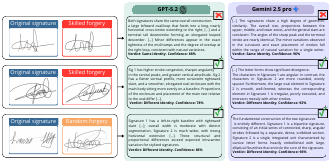} 
    
    \caption{\textbf{Qualitative Exploration of VLM Forensic Reasoning.} We analyze three distinct scenarios from the SVC evaluation dataset. 
    \textbf{Top row - The Rationalization Trap:} In a challenging skilled forgery, both models fail (False Positive), hallucinating that the visible kinematic defects are natural variations. 
    \textbf{Middle row - True Detection:} A successful skilled forgery detection where the models correctly identify subtle differences in ``stroke congestion" and ``angularity" (True Negative). 
    \textbf{Bottom row - Topological Reasoning:} A random forgery scenario where the models act as perfect geometric discriminators, identifying the ``bipartite" vs. ``integrated" structure.}
    \label{fig:qualitative_analysis}
\end{figure*}

Skilled forgeries present a morphologically congruent shape but differ in kinematic execution. We observe two distinct behaviors in this scenario: 1) Rationalization Trap (Failure): The top row of Figure \ref{fig:qualitative_analysis} illustrates a critical failure. Visually, the forgery exhibits signs of a slow-hand imitation, lacking firmness. However, the VLMs prioritize the geometric match over the line quality. Instead of identifying the tremor as deceit. GPT-5.2 dismisses the artifacts as ``consistent with natural variation or generative jitter", effectively creating a narrative that forgives the biometric flaws. Similarly, Gemini asserts that minor curvature variations ``fall within the range of natural variation". This exposes another severe failure: despite the visual ambiguity, both GPT and Gemini assign a high confidence (88\% and 92\%, respectively) to this error. 2) The Success Detection: The middle row of Figure \ref{fig:qualitative_analysis} demonstrates that VLMs can detect skilled forgeries when distinct stylistic markers are present. Here, GPT correctly identifies ``higher stroke congestion" and ``sharper angularity" in the genuine sample compared to the ``smoother" forgery. Gemini also correctly distinguishes the ``angular, jagged" characters of the original from the ``rounded, evenly spaced" impostor. This suggests that while VLMs struggle with subtle dynamics, they can remain effective at detecting stylistic divergences in stroke formation.

\vspace{-2mm}

\section{Conclusion}
\label{sec:conclusion}
This study constitutes an initial exploration of the use of the two leading commercial VLMs for zero-shot signature verification. The findings indicate a dichotomy in performance, whereby the models demonstrate remarkable robustness in random comparisons; however, their reliability decreases in the context of skilled forgeries, where there is an attempt to rationalize stroke artifacts as natural variations through CoT reasoning. Due to these inconsistencies, the viability of VLMs for autonomous operation in the forensic field is currently limited. However, their potential as explainability assistants for human experts is promising. During the course of our research, we detected the presence of recurring cinematic hallucinations, which we intend to examine in detail as future work.  We will also focus on the detailed analysis of the explanations generated, the evaluation of open-source models and the implementation of few-shot learning strategies. 

\section*{Acknowledgement}
This project has been supported by PowerAI+ (SI4/PJI/2024- 00062 Comunidad de Madrid and UAM), Cátedra ENIA UAM-Veridas en IA Responsable (NextGenerationEU PRTR TSI-100927-2023-2), and TRUST-ID (PID2025-173396OB-I00 MICIU/AEI and the EU). Robledo-Moreno is supported by a FPI Fellowship (FPI-UAM-2025). Also, we acknowledge the computer resources provided by Centro de Computación Científica-UAM.

\vspace{2mm}

\bibliographystyle{IEEEtran}
\bibliography{main}

\end{document}